\documentclass[runningheads]{llncs}
\usepackage[T1]{fontenc}
\usepackage{graphicx}
\usepackage{booktabs}
\usepackage{tabularx}
\usepackage{multirow}
\usepackage{multicol}
\usepackage{array}
\usepackage{float}
\usepackage{subcaption}

\usepackage{amsmath, amssymb, amsfonts}
\usepackage{textcomp}

\usepackage{xcolor, soul}
\usepackage{caption}
\usepackage{setspace}
\usepackage{enumitem}
\usepackage{placeins}

\usepackage{url}
\usepackage{cite}

\usepackage{soul}
\begin{document}
\title{Detecting and Understanding Hateful Contents in Memes Through Captioning and Visual Question-Answering}
\titlerunning{Detecting and Understanding Hateful Contents in Memes}
%
%
 \author{
 Ali Anaissi\inst{1,3} \and
Junaid Akram \inst{1,3,4} \and
Kunal Chaturvedi\inst{2} \and
 Ali Braytee\inst{2}}
 \authorrunning{A. Anaissi et al.}
\institute{The University of Sydney, School of Computer Science, Camperdown, NSW 2008, Australia
\email{junaid.akram@sydney.edu.au, ali.anaissi@sydney.edu.au} \and
University of Technology Sydney, School of Computer Science, Ultimo, Australia\\
\email{Kunal.Chaturvedi@uts.edu.au, ali.braytee@uts.edu.au} \and
University of Technology Sydney, TD School, Ultimo, Australia\\
\email{ali.anaissi@uts.edu.au, junaid.akram@uts.edu.au} \and
Australian Catholic University, Peter Faber Business School, North Sydney, NSW 2060 Australia \\
\email{junaid.akram@acu.edu.au}
}

\maketitle              
\begin{abstract}
Memes are widely used for humor and cultural commentary, but they are increasingly exploited to spread hateful content. Due to their multimodal nature, hateful memes often evade traditional text-only or image-only detection systems, particularly when they employ subtle or coded references. To address these challenges, we propose a multimodal hate detection framework that integrates key components: OCR to extract embedded text, captioning to describe visual content neutrally, sub-label classification for granular categorization of hateful content, RAG for contextually relevant retrieval, and VQA for iterative analysis of symbolic and contextual cues. This enables the framework to uncover latent signals that simpler pipelines fail to detect. Experimental results on the Facebook Hateful Memes dataset reveal that the proposed framework exceeds the performance of unimodal and conventional multimodal models in both accuracy and AUC-ROC. 

\end{abstract}
\section{Introduction}

Memes have emerged as a widely used medium on social media platforms, combining images and text overlays to convey humor, satire, or cultural commentary. Despite their seemingly innocuous appearance, memes are increasingly exploited to propagate hateful or discriminatory content \cite{munzni2024classification,akram2018lexicon}. Due to their multimodal nature, such content often bypasses conventional text-only or image-only detection algorithms. When image elements and textual components interact in subtle ways, hateful content may remain hidden, allowing offending material to circulate unchecked\cite{akram2024galtrust, blaier2021caption}. Empirical evidence from the Hateful Memes Challenge has shown that unimodal approaches typically fail to adequately capture the range of possible hateful expressions embedded within memes \cite{HatefulMemesChallenge,kirk2021memeswild}. Consequently, there is a critical demand for robust, integrated solutions that can parse both textual and visual cues to identify underlying animosity or prejudice.

Recent studies \cite{10.1145/3485447.3512260, hamza2023multimodal,liu2024ensemble} have attempted to bridge the gap between language and vision representations, revealing that combined multimodal strategies can achieve promising results for specific domains such as misogynistic memes \cite{rizzi2023recognizing} or harmful COVID-19 memes \cite{pramanick-etal-2021-detecting}. While these approaches have shown promise, they exhibit several key limitations that hinder their ability to comprehensively detect nuanced hateful content. First, many existing methods \cite{10.1007/978-3-319-93417-4_48, anaissi2024fine,akram2024ai} rely on fixed multimodal representations, where text and image features are extracted independently and fused statically. This rigid approach fails to capture the dynamic interplay between textual and visual cues, making it difficult to detect contextually embedded hate signals, such as sarcasm, coded symbols, or ambiguous imagery \cite{kougia2021multimodal}. Second, these methods typically lack real-time adaptive reasoning, instead relying on predefined classification heuristics \cite{kovacs_challenges_2021,rehman2018statistical}. As a result, they struggle with detecting veiled or evolving hate speech that requires contextual reasoning beyond surface-level analysis. Third, existing models often categorize hateful content using coarse-grained labels, such as simply hateful or non-hateful, without distinguishing between different forms of hate speech. This lack of specificity reduces interpretability and makes it harder to apply targeted moderation strategies. Such limitations highlight the need for a more systematic approach that incorporates iterative questioning, refined retrieval, and text-image fusion at a granular level.

To address these challenges, this paper introduces a framework that integrates optical character recognition (OCR), caption generation, retrieval-augmented classification, and a visual question answering (VQA) module. OCR reliably extracts overlaid text, while captioning supplies a neutral description of the visual scene. We enhance classification by leveraging a sub-labeling strategy, segmenting hateful content according to attributes such as race, religion, or others. This fine-grained division increases precision in retrieval-augmented steps, ensuring that exemplars align more closely with the observed meme. Additionally, the VQA system formulates targeted queries about potentially harmful symbols, background contexts, or linguistic cues that might escape notice in single-round analyses. By integrating these components, we aim to offer a system robust enough to detect concealed instances of hate speech.


The paper makes the following contributions:
\begin{itemize}
\item A multimodal approach that integrates OCR for textual extraction, neutral captioning for visual context, a sub-label retrieval, and a multi-turn VQA, to detect both explicit and implicit hateful cues in memes is proposed. 


\item A sub-label classification framework that partitions hateful content into race-based, gender-based, and other sub-dimensions is introduced to improve the accuracy of retrieval-augmented generation (RAG). 





\end{itemize}

\section{Related Works}

Over the past few years, research on hateful meme detection has evolved considerably, emphasizing the need for integrated analysis of both textual and visual modalities\cite{qian2024optimized,khan2021spice}. Early attempts often separated images from text, applying standard classifiers to each modality in isolation. However, the limitation of such methods became apparent when memes contained subtle or implicit hateful references that only emerged through interaction between visual features and overlaid text. Consequently, various studies started to explore multimodal fusion. Kiela et al. \cite{kiela2021hatefulmemes} introduced the Hateful Memes Challenge, releasing a dataset that paired each image with short textual content to highlight the complexities of meme-based hate. Badjatiya et al. \cite{badjatiya2017deep} and Davidson et al. \cite{davidson2017automated} initially concentrated on textual classification, adopting lexicon-based approaches or neural architectures like CNNs and LSTMs, but these did not fully capture the compound nature of memes. Meanwhile, image-based methods such as Gómez et al. \cite{gomez2020exploring} and Howard et al. \cite{howard2019searching} attempted to detect hateful symbols or cues through CNNs and other vision models, yet struggled when the hatred was expressed solely via text.

Subsequent efforts introduced hybrid or multimodal models to process images and text jointly. Transformative architectures such as ViLBERT \cite{lu2019vilbert} and Visual BERT \cite{li2019visualbertsimpleperformantbaseline} harness cross-attention mechanisms to align textual and visual embeddings, thereby improving classification accuracy. In parallel, the Facebook Hateful Memes dataset \cite{kiela2021hatefulmemes} further prompted researchers to refine their multimodal pipelines, as it contained nuanced and challenging examples of encoded hate speech. Rizzi et al. \cite{rizzi2023recognizing} addressed misogynistic memes by proposing a fine-tuned VisualBERT that excelled at combining textual embeddings from OCR with high-level image features obtained from pretrained CNNs. Pramanick et al. \cite{pramanick-etal-2021-detecting} tackled COVID-19-related misinformation with a focus on harmful memes, showing that domain-specific training data could refine detection for medical or pandemic-oriented hate. Although these approaches outperformed unimodal baselines, they occasionally failed on memes whose meaning shifted dramatically depending on cultural or contextual details not captured by purely data-driven models.

Recent works have sought to incorporate advanced language models, retrieval techniques, and Visual Question Answering (VQA) to overcome the remaining challenges. Devlin et al. \cite{devlin2019bertpretrainingdeepbidirectional} illustrated the utility of contextual embeddings via BERT for language understanding, while Lewis et al. \cite{lewis2020rag} introduced Retrieval-Augmented Generation (RAG) to infuse external knowledge into classification or generation tasks. Accordingly, sub-label methods emerged to partition hateful content into categories such as race or gender, facilitating more precise retrieval and classification \cite{rizzi2023recognizing}. Additionally, VQA-based systems proved beneficial for generating iterative queries about scene elements, as multi-turn dialogue can reveal latent meaning. By incorporating refined embedding models like CLIP \cite{radford2021learning} and advanced prompt engineering, researchers succeeded in capturing the interplay between textual overlays and visual symbolism at deeper levels  \cite{hamza2023multimodal,agarwal2024mememqa}. Collectively, these investigations underscore the vital role of multimodality and contextual verification in tackling hateful memes, thereby guiding the development of more robust pipelines that can identify concealed or culturally coded hatred.

\section{Methods}

\begin{figure}[t] 
\centering 
\includegraphics[width=1\linewidth]{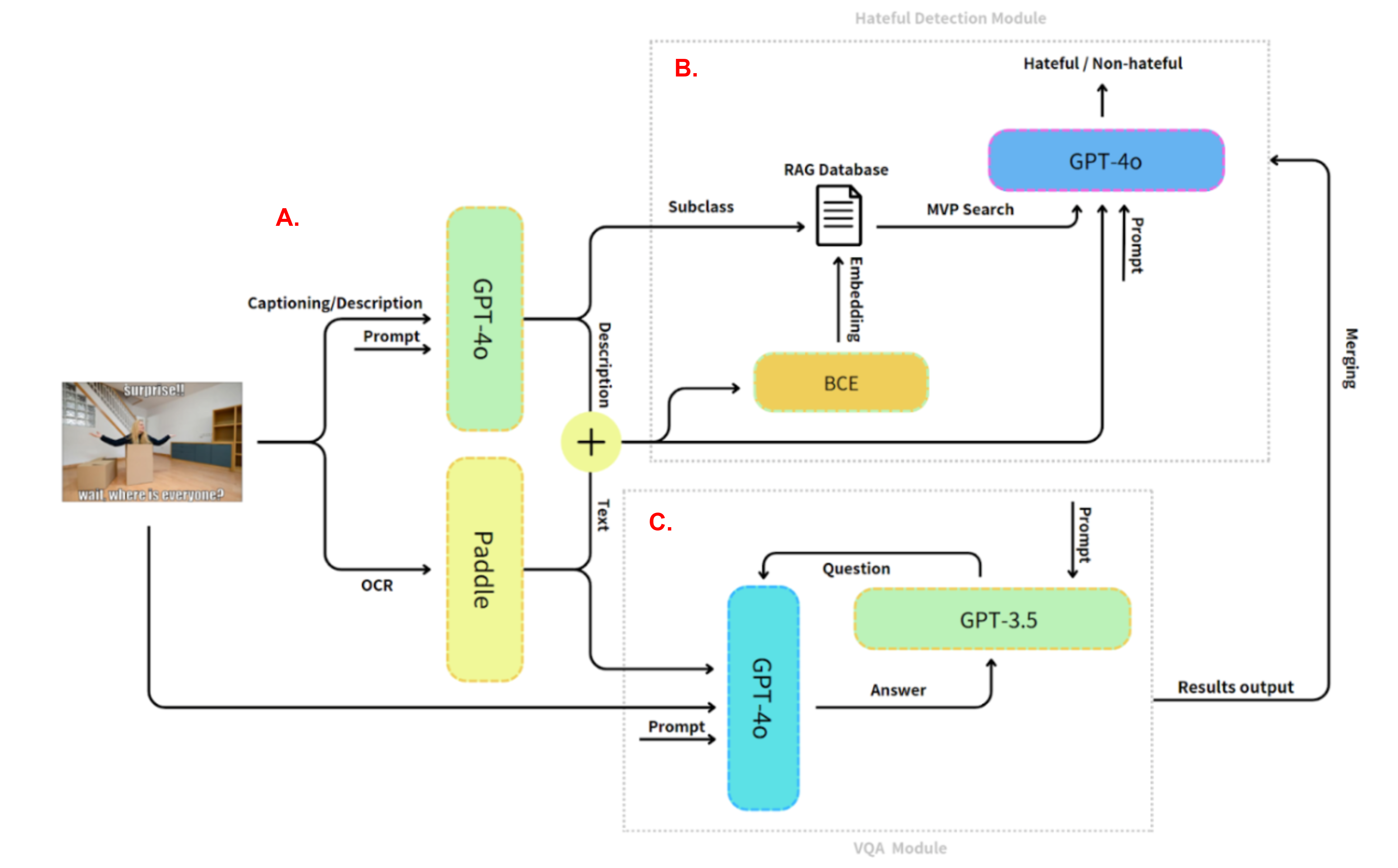} \caption{The overall framework, RAG (sub\_label + VQA) for detecting hateful content} \label{fig:completeframe} 
\end{figure}

In this section, we describe our proposed multimodal pipeline integrating OCR and captioning, VQA module, and a hateful detection module. The overall framework is shown in Fig. \ref{fig:completeframe}.

\subsection{Captioning and OCR}
A key challenge in detecting hateful content in memes arises from the interplay of textual and visual cues. As shown in Fig. \ref{fig:completeframe}A, we first extract all text and generate captions before passing the multimodal information to our detection modules. Specifically, we separate the input processing into two complementary procedures. First, Paddle OCR \cite{du2020ppocrpracticalultralightweight}, an optical character recognition system, is used to extract textual messages in memes. To supplement OCR, we generate a caption describing the visual content of the meme using a large language model. The net effect is a more robust representation of the meme, combining the recognized text with a broad contextual description.

\subsection{VQA Module}

The Visual Question Answering module integrates OCR, multimodal analysis, and advanced language models to detect and analyze hateful content in memes. As shown in Fig. \ref{fig:completeframe}C, the workflow begins with raw meme input, which includes both visual and textual data. Using the Paddle OCR module, textual information embedded within the image is extracted, enabling the identification of captions, phrases, or symbols that may carry hateful messages. This extracted text, along with the raw image, forms the foundation for multimodal analysis in the VQA pipeline. The second stage involves processing the inputs using GPT-based models for dynamic question generation and context-aware answering. Initially, GPT-4.0 generates a broad, context-sensitive question aimed at understanding the overall theme of the meme, with a focus on detecting hate signals such as stereotypes, offensive language, or harmful visual elements. The generated question is then processed by GPT-3.5, which provides detailed answers by integrating visual and textual cues. If hate-related elements are identified, such as racial slurs or stereotypical imagery, the system refines its analysis by generating follow-up questions through GPT-4.0, specifically targeting the hateful components.

To ensure coherence and accuracy, the system employs a multi-turn dialogue mechanism, where all questions and answers are stored in a contextual database. This enables the system to maintain continuity across interactions, eliminating redundancy and ensuring that every aspect of the meme is thoroughly examined. Once the analysis is complete, the structured Q\&A pairs are fed into a specialized Hate Detection Module that works in conjunction with the Retrieval-Augmented Generation (RAG) pipeline. The RAG pipeline further contextualizes and validates the findings by cross-referencing against a knowledge base of hate-related symbols, phrases, and behaviors. The final output of the system includes a detailed summary of the detected hateful content, highlighting specific textual elements, visual cues, and their contextual implications. By integrating OCR, multimodal analysis, and dynamic reasoning, the VQA system provides a robust solution for detecting hateful content in memes. Its iterative question-answering logic, combined with adaptive refinement, ensures thorough exploration of nuanced hate signals, making it a powerful tool for content moderation in real-world scenarios.

\subsection{Hateful Detection}

In the hateful detection module, we aim to classify whether a meme is hateful or non-hateful by leveraging the OCR text, the generated caption, and outputs from the VQA module. Fig. \ref{fig:completeframe}B illustrates the schematic of our hateful content detection module, which consists of multiple steps, including RAG, sub-labeling, and content explanation..

\textbf{RAG Overview: } The Retrieval-Augmented Generation (RAG) architecture~\cite{lewis2020rag} is utilized as a core component for hate detection by incorporating a vector database, embedding model, and ranking mechanism. The architecture processes each meme’s textual input (caption and OCR) by embedding it and querying a repository of labeled data, explanations, and examples. These retrieved chunks provide additional context in a prompt-like fashion to guide the model in producing more accurate and context-aligned inferences about hateful content. 

\textbf{Content Explanation for RAG: } Another variant of the RAG architecture augmented the vector database with not only the caption and OCR text but also detailed explanations for why specific memes were labeled hateful or non-hateful. The assumption was that these explanations could help the model identify nuanced hate signals in new memes. 

\textbf{Sub-labeling for RAG: } A specific implementation of RAG, known as “sub-label classification”, was applied. Instead of treating hatefulness as a single, broad category, the sub-labeling method divides hateful content into finer-grained categories, such as race, religion, and others. By embedding the meme’s caption and OCR text, the RAG system retrieves content related to the most relevant sub-label, providing contextual anchors that improve classification. For instance, a meme involving race-based hate speech retrieves examples and contextual references from the race sub-label category, making the detection more precise.

Our final framework, RAG (sub\_label + VQA) integrated the outputs of the VQA module with the sub-labeling RAG method. The VQA results, which provide detailed contextual information by combining caption and OCR analysis with visual cues, were incorporated into the RAG pipeline as additional reference material to detect hateful memes. 




\section{Experiments and Results}


\subsection{Datasets}

We utilized Facebook Hateful Memes (FHM) \cite{HatefulMemesChallenge} dataset for our experiments. The dataset contains diverse meme examples with hateful vs. non-hateful labels. The data integrates text captions overlaid on images and is one of the primary resources provided by Facebook for the Hateful Memes Challenge. 
Next, we apply random transformations such as rotation, scaling, cropping, and mild color jitter. These transformations enrich the model’s exposure to diverse visual conditions, thus boosting resilience to typical noise or distortion in user-generated memes. In certain data splits, we note that hateful content is encoded through metaphors, coded language, or domain-specific references. We thus expand the dataset with carefully curated examples reflecting these nuances to reinforce the sub-labeling strategy in the RAG pipeline. This expansion aids in capturing cultural, linguistic, or other contextual factors that might not be evident from standard data subsets. Overall, these strategies enhance the system’s capacity to tackle newly emerging hateful memes with novel textual or visual patterns.

\subsection{Evaluation Metrics}

For hateful content detection, we employ two primary metrics: Accuracy and AUC-ROC. For VQA, we adopt the VQAScore methodology \cite{Lin2024Evaluating}. 
We conducted five rounds of scoring to mitigate model variability. In each round, the VQA system generated answers to a collection of queries derived from the meme images. We then calculated VQAScore for each generated answer-image pair, and took the average over these five rounds. This approach ensures a more robust estimation of performance, reducing the influence of any single outlier run.

\begin{table}
\caption{Performance comparison across various models, including unimodal and multimodal methods. Acc. denotes Accuracy; AUROC stands for Area Under the Receiver Operating Characteristic.}
\label{tab:finalresults}
\centering
\begin{tabular}{llcc}
\hline
\textbf{Type} & \textbf{Model} & \textbf{Acc. (\%)} & \textbf{AUROC (\%)} \\
\hline
 & Human annotators & 84.70 & 82.65 \\
\hline
\multirow{3}{*}{Unimodal} 
 & Image-grid \cite{he2015deepresiduallearningimage} & 52.00 & 52.63 \\
 & Image-region \cite{kiela2021hatefulmemes} & 52.13 & 55.92 \\
 & Text BERT \cite{devlin2019bertpretrainingdeepbidirectional} & 59.20 & 65.08 \\
\hline
\multirow{9}{*}{Multimodal} 
 & Late fusion \cite{kiela2021hatefulmemes} & 59.66 & 64.75 \\
 & Concat BERT \cite{kiela2021hatefulmemes} & 59.13 & 65.79 \\
 & MMBT-grid \cite{kiela2021hatefulmemes} & 60.06 & 67.92 \\
 & MMBT-region \cite{kiela2021hatefulmemes} & 60.23 & 70.73 \\
 & ViLBERT \cite{lu2019vilbertpretrainingtaskagnosticvisiolinguistic} & 62.30 & 70.45 \\
 & Visual BERT \cite{li2019visualbertsimpleperformantbaseline} & 63.20 & 71.33 \\
 & ViLBERT CC \cite{kiela2021hatefulmemes} & 61.10 & 70.03 \\
 & Visual BERT COCO \cite{kiela2021hatefulmemes} & 64.73 & 71.41 \\
 & GPT-4o mini \cite{openai2024gpt4o} & 69.50 & 75.02 \\
\hline
\multirow{3}{*}{\textbf{Our Method}} 
 & RAG (explanation) & 59.20 & 63.01 \\
 & RAG (sub\_label) & 72.00 & 76.52 \\
 & RAG (sub\_label + VQA) & \textbf{73.50} & \textbf{78.35} \\
\hline
\end{tabular}
\end{table}

\subsection{Results}

\subsubsection{Quantitative Analysis}

Table~\ref{tab:finalresults} summarizes the performance of various models and methods in detecting hateful memes. The table also includes comparisons with human annotations as an upper bound, as well as benchmark approaches from the challenge. Human annotations remain the most accurate, with 84.70\% accuracy and 82.65\% AUROC. Among unimodal methods, vision-only models such as Image-grid and Image-region perform poorly around 52\% accuracy, highlighting the insufficiency of visual cues alone for detecting nuanced hateful content. Text BERT outperforms these with 59.20\% accuracy and 65.08\% AUROC, underscoring the greater informativeness of textual features in this domain. Multimodal baselines, which integrate image and text modalities, demonstrate marked improvements. Simple fusion techniques like Late Fusion and Concat BERT offer modest gains (~59–60\% accuracy). More sophisticated architectures such as MMBT-Region, ViLBERT, and Visual BERT COCO further improve performance, reaching up to 64.73\% accuracy and 71.41\% AUROC. The strongest baseline, GPT-4o mini, achieves 69.50\% accuracy and 75.02\% AUROC, setting a high bar for general-purpose large multimodal models.

Our proposed method significantly outperforms all baselines. While the RAG (explanation) variant performs comparably to Text BERT with 59.20\% accuracy, incorporating fine-grained sub-labels in RAG leads to a substantial boost with accuracy of 72.00\% and AUROC of 76.52\%. The highest gains are observed when this sub-label retrieval is further combined with VQA, yielding the best overall results of 73.50\% accuracy and 78.35\% AUROC. These findings confirm the synergy between sub\_label-based retrieval and the contextual enhancements provided by the VQA module. Rather than only relying on raw text or naive retrieval from explanation templates, the sub\_label approach retrieves precisely relevant hateful exemplars, while the VQA module helps uncover implicit cues that might not be evident through OCR captioning alone.



\subsubsection{Qualitative Observations}

\begin{figure}
    \centering
    \includegraphics[width=\linewidth]{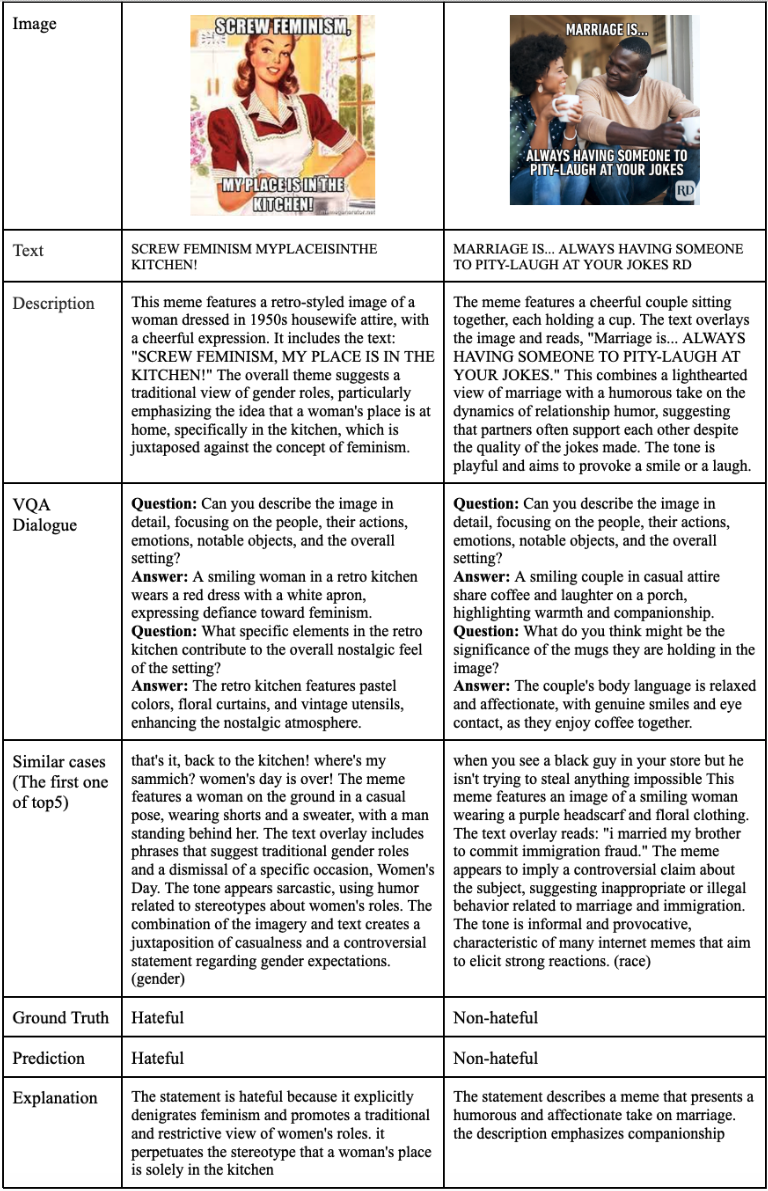}
    \caption{Outputs for a hateful example (left) and a non-hateful example (right). The pipeline includes accurate OCR detection, objective captioning, multi-turn VQA addressing targeted hate cues, and final classification via RAG.}
    \label{fig:enter-label}
\end{figure}

Fig. \ref{fig:enter-label} offers an illustrative example, showing system outputs for both a positively identified hateful meme and a non-hateful instance. The figure includes how OCR extracts textual content, how the captioning module describes the image, and how the multi-round VQA interacts with the meme to highlight potentially hateful elements.

In the hateful example, OCR precisely captured key terms from the overlaid text, and the captioning module accurately noted contextual objects and background. The VQA dialog then focused on potentially discriminatory language, confirming hateful cues and retrieving relevant sub\_label data through the RAG sub\_label pipeline. The final classification was correct and accompanied by a short textual explanation consistent with the known ground truth.

In the non-hateful case, the system again accurately recognized textual and visual details but found no hateful signals. The RAG retrieval was less relevant, returning only examples bearing minimal resemblance to hateful content. As a result, the classification was non-hateful, aligning with the ground-truth label. This outcome underlines the pipeline’s capacity to remain conservative when the textual and visual signals do not suggest hateful references.

\subsection{Discussion}

These results indicate that combining large language models with sub\_label-based retrieval and VQA modules yields significant gains over unimodal or simpler multimodal baselines. Several observations are worth highlighting:

\textbf{Performance gaps and remaining challenges.} Although RAG (sub\_label) + VQA achieves 73.50\% accuracy, it still lags behind human annotators. This gap underscores the complexity and ambiguity of hateful memes, which often rely on cultural references, double meanings, or evolving slang not always captured by static training data. Further refinement of sub\_label categories, addition of external knowledge sources, and extended data augmentation strategies may narrow this human-machine divide.

\textbf{Effectiveness of OCR} With the rise of multimodal large language models, one may question the need for a dedicated OCR module. However, we find that incorporating explicit OCR (PaddleOCR) remains valuable, especially when dealing with stylized, distorted, or meme-specific fonts that challenge even state-of-the-art vision-language models. Explicit OCR ensures consistent and controllable extraction of embedded text, which downstream modules such as VQA and RAG rely on for accurate reasoning. Furthermore, separating text extraction from high-level reasoning supports interpretability, and modular debugging.

\textbf{Utility of VQA dialogue.} Notably, RAG showed visible improvements after including VQA-derived context. The VQA system probes the meme with targeted questions, clarifying ambiguous cues and capturing nuanced correlations between text and visuals. This synergy is crucial in uncovering content that is hateful only when certain textual or symbolic aspects align with specific contexts or objects in the image. The average VQAScore of 75.04 also suggests that the system reliably produces answers consistent with the underlying image content, thereby strengthening the subsequent classification. 

\textbf{Explanation-based RAG limitations.} The RAG (explanation) configuration performed poorly for reasons related to noise in the textual explanations and potential misalignment with new memes. The assumption that labeled explanations from certain memes would be directly transferable or consistently interpreted appears flawed. In contrast, sub\_label retrieval provides a more targeted anchor (e.g., detecting “racial hate” or “religious hate” specifically), improving retrieval precision.


\textbf{Implications for real-world applications.} Content moderation platforms or social media sites that must identify hateful memes in real time can benefit from adopting a pipeline that integrates a carefully designed retrieval mechanism, a multi-turn VQA system, and robust text-image analysis. However, real-time constraints require optimization to reduce computational overhead; our solution underscores the effectiveness of multi-step synergy but also reveals potential latency in large-scale deployments. In particular, the use of large language models for multi-turn VQA and the dependency on retrieval-augmented generation (RAG) with sub-label classification introduce substantial memory and processing demands. These may hinder responsiveness in high-throughput or latency-sensitive settings. Future engineering efforts would thus focus on accelerating sub\_label lookups and streamlining the VQA query-response phase.

The quantitative results, shown in Table~\ref{tab:finalresults}, and the qualitative analysis demonstrate that our integrated system effectively detects hateful memes and outperforms several multimodal baselines. Although there remains a gap relative to human-level comprehension of subtle, context-dependent hate, the results confirm that careful synergy among OCR, captioning, VQA, and specialized retrieval strategies can significantly improve classification performance. 


\section{Conclusion}
The proposed framework demonstrates a robust approach to addressing hateful content detection within memes by integrating multiple modules for text extraction, captioning, retrieval, and visual question answering. The integrated pipeline achieves significant accuracy and AUC-ROC compared to existing methods on established benchmarks. These results underline the importance of uniting refined language strategies with methods that analyze images more deeply. Future work could extend the framework by incorporating culturally nuanced knowledge graphs, refining VQA prompts to reduce false positives, or integrating dynamic feedback loops for real-time detection. 




\section*{Acknowledgment}
We acknowledge the contributions of Bonnie Zhong, Haodi Yang, Yinuo Wang, Lu Tang, Yiqi Zhao, Yuanhao Huo to this project.

\bibliographystyle{splncs04}
\bibliography{sample-base}
\end{document}